\documentclass[11pt,a4paper]{article}
\usepackage[hyperref]{nlp4MusA}
\usepackage{times}
\usepackage{url}
\usepackage{amssymb}
\usepackage{xcolor}
\usepackage{latexsym}
\usepackage{subfig}
\usepackage{varwidth}
\usepackage{graphicx}
\usepackage{cases}
\usepackage{placeins}
\usepackage{array, boldline, makecell, booktabs}
\usepackage{multirow}
\usepackage{multicol}
\newcommand{\comment}[1]{}

\aclfinalcopy % Uncomment this line for the final submission
\newcommand\BibTeX{B\textsc{ib}\TeX}

\title{Contextualized Spoken Word Representations Using Convolutional Autoencoders}

\author{Prakamya Mishra \\
  Shiv Nadar University \\
  \texttt{pm669@snu.edu.in} \\\And
  Pranav Mathur \\
  Shiv Nadar University \\
  \texttt{pm872@snu.edu.in} \\}

\date{}

\begin{document}
\maketitle
\begin{abstract}
% A lot of work has been done recently to build text-based language models, but not much has been done to model speech/audio. In the case of text, words are represented by unique fixed-length embeddings that hold semantical and syntactic relationships between words. Such embeddings for speech/audio type data can not only lead to great advances in the speech/audio related natural language processing tasks but can also reduce the loss of information like tone, expression, accent, etc while converting speech to text in order to perform these tasks. This paper proposes a novel model architecture that produces syntactically and semantically adequate contextualized representation of varying length spoken words. The performance of the spoken word embeddings generated by the proposed model was validated by (1) inspecting the vector space generated, and (2) evaluating its performance on three benchmark datasets for measuring word similarities.

A lot of work has been done to build text-based language models for performing different NLP tasks, but not much research has been done in the case of audio-based language models. This paper proposes a Convolutional Autoencoder based neural architecture to model syntactically and semantically adequate contextualized representations of varying length spoken words. The use of such representations can not only lead to great advances in the audio-based NLP tasks but can also curtail the loss of information like tone, expression, accent, etc while converting speech to text to perform these tasks. The performance of the proposed model is validated by (1) examining the generated vector space, and (2) evaluating its performance on three benchmark datasets for measuring word similarities, against existing widely used text-based language models that are trained on the transcriptions. The proposed model was able to demonstrate its robustness when compared to the other two language-based models.

\end{abstract}

\section{Introduction}
There are several methods in which humans and computers can converse, like speaking (audio) and writing (text). At present, research in the field of NLP has advanced a lot to attain a good understanding of textual data but there are still some ways to go to properly contemplate the audio/speech data. 

Word embeddings are extensively used in NLP applications since they have proven to be an extremely informative representation of the textual data. Language models like GloVe \cite{pennington-etal-2014-glove} and Word2Vec \cite{NIPS2013_5021} successfully transform textual words from its raw form to semantically and syntactically correct, fixed dimensional vectors. These type of word representations for the spoken words can be widely used to process speech/audio data for tasks like Automatic Summarization {\cite{kageback-etal-2014-extractive}}, Machine Translation {\cite{DBLP:journals/corr/Jansen17a}}, Named Entity Recognition {\cite{10.1007/978-981-13-9409-6_218}}, Sentiment Analysis {\cite{DBLP:journals/corr/Liu17b}}, Information Retrieval {\cite{DBLP:journals/corr/RekabsazMLH17}}, Speech Recognition {\cite{DBLP:journals/corr/abs-1902-06833}}, Question-Answering {\cite{Tapaswi_2016_CVPR}} etc.

% The Word2Vec model {\cite{NIPS2013_5021}} successfully transforms textual words from its raw form to semantically and syntactically correct, fixed dimensional vectors.
Compared to text, not much research has been done in the field of audio-based modeling primarily due to the lack of large, reliable, clean, and publicly available datasets on which the spoken word language models can be trained. Also, spoken words unlike textual words have a different meaning when they are spoken in a different tone, expression, accent, etc, and incorporating them exponentially increases the difficulty of building such language models. Such models also face difficulties such as different people can have different pronunciations, tones, and pauses for the exact same words.

The proposed model, aims at generating syntactically and semantically adequate contextualized vector representation of the variable length audio files (instead of using fixed length audio files with multiple word utterances), where each file corresponds to a single spoken word in a speech and further validates the vector representations by evaluating it on three benchmark word similarity datasets (SimVerb, WS-SIM, WS-REL). To further increase the interpretability, this paper also provides illustrations of the vector space generated by the proposed model.

\section{Related Work}
A lot of work has been done in the field of NLP to give textual words sound representations. Word2Vec \cite{NIPS2013_5021} has demonstrated huge improvements in embedding sub-linear relationships into the vector space of the words but at the same time, they were unable to handle out of vocabulary words. Another comparable word representation model is GloVe \cite{pennington-etal-2014-glove}. GloVe works to fit a giant word co-occurrence matrix built from the matrix. GloVe helps in taking into account the semantics and also gives relatively smaller dimension vectors.

Recent advances have enabled it to apply deep learning to transform spoken word segments into fixed dimensional vectors. {\cite{chung2016audio}}, uses fixed-length audio files and passes them through a Sequence-to-Sequence Autoencoder (SA) and Denoising SA (DSA) to generate word embeddings. They demonstrated that the phonetically similar words had close spatial representations in the vector space but they failed to meet the result standards similar to those by GloVe trained on Wikipedia. Following the above work, {\cite{chung2017learning}} used 500 hours of speeches from multiple speakers divided into fixed audio segments. They compare the results also with GloVe based on 13 different comparison measures. Both, {\cite{chung2016audio}} \& {\cite{chung2017learning}}  failed to capture the spoken words properly due to the use of fixed length audio segments. {\newcite{9060816}}, proposed an audio2vec model which was built on top of the Word2Vec models (Skip-gram \& CBOW) to reconstruct spectrogram slices using the contextual slices and temporal gaps. They were able to show that Audio2Vec performed better than the other existing fully-supervised models.

\section{Model}

\begin{figure*}[ht] % picture
    \centering
    \includegraphics[width=0.7\textwidth]{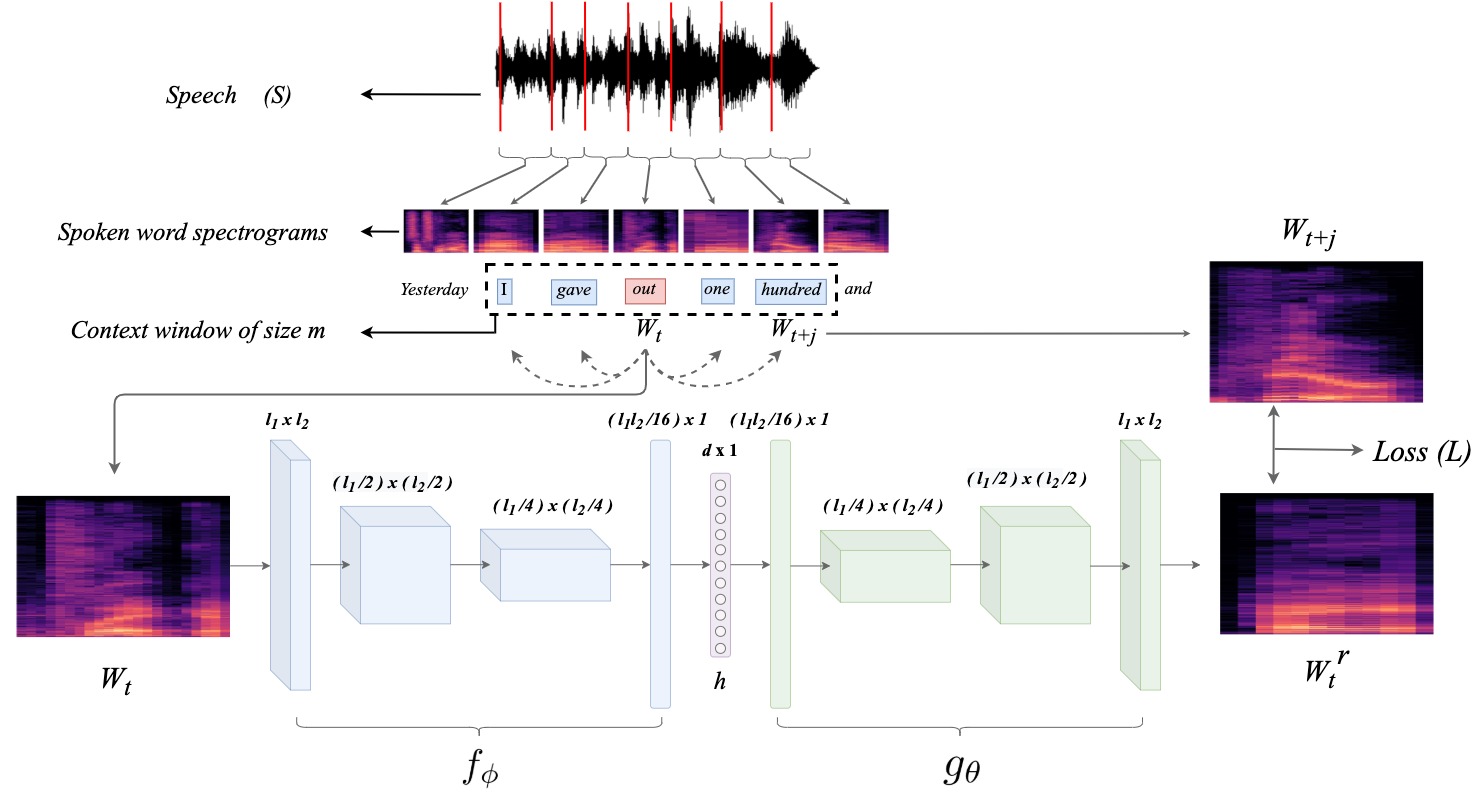}
    \caption{\small{Proposed Model Architecture}}
    \label{fig:main_model}
\end{figure*}

The proposed model uses sequential utterances of words from a speech to learn their corresponding contextualized representations. These learned contextualized representations capture the semantic and syntactic properties of these spoken words. The input to the model is a speech \(S\). This speech is split into individual spoken word utterances (independent variable-length audio files). The proposed model used audio spectrograms for representing the audio files of these spoken word utterances. An audio spectrogram is a visual representation of sound. So to get spectral representations, all the spoken word utterances are converted to their corresponding spectrograms (which depicts the spectral density of a sound \textit{w.r.t} time (in our case an utterance)). The spoken word utterance spectrograms are represented by \(W_i\) as shown in equation \ref{equ1}.
\begin{equation} \label{equ1}
S = [W_1, W_2, . . ., W_n], n \in \mathbb{R}
\end{equation}
In the above equation \(n\) represents the total number of spoken words present in a sentence of the speech and \(W_i \in \mathbb{R}^{{l_1}{l_2}}\) represents a spectrogram, where, \(l_1\) is for the frequency (pitch/tone) dimension, \(l_2\)  represents time. Values in the spectrogram represents amplitude (energy/loudness) at a particular time of a particular frequency.

Words have different meanings when they are spoken in different contexts. To capture the context corresponding to spoken words, the proposed model uses a context window of size \(m\). So the representation of a spoken word (target word) is learned based on \(m\) spoken words after and before it. This context window of size \(m\) slides over the whole speech having a target spoken word \(W_t\) (where \(1 \leq t \leq n\)) at the middle and \(m\) context spoken words before and after it (a total of 2m context words). These context spoken words are represented by \(W_{t+j}\) where \(-m \leq j \leq m\) \& \(j \neq 0\).

Next, the model passes all the pairs of the target spoken word spectrograms \(W_t\) with its corresponding context spoken word spectrograms \(W_{t+j}\) into a convolutional autoencoder individually to learn the contextual representation of the target spoken word corresponding to \(W_t\). The convolutional autoencoder is composed of two independent neural networks namely, an encoder network and a decoder network. The encoder network is represented by \(f_{\phi}\) and the decoder network is represented by \(g_{\theta}\), where \(\phi\) and \(\theta\) are the learnable parameters corresponding to both the networks. Both \(f_{\phi}\) \& \(g_{\theta}\) are used to extract the spatial features of the input spectrogram \textit{w.r.t} to the output spectrogram. The target spoken word spectrogram \(W_t\) is given as input to the encoder network, which outputs a latent representation \(h\). This latent representation is then given as input to the decoder network, i.e.

\begin{equation} \label{equ2}
\small{h = f_{\phi}(W_t) = \sigma(W_t \ast \phi)}
\end{equation}
\begin{equation} \label{equ3}
\small{W_{t+j} = g_{\theta}(h) = \sigma(h \ast \theta)}
\end{equation}
\begin{equation} \label{equ4}
\small{W_{t+j} = g_{\theta}(f_{\phi}(W_t))}
\end{equation}

In the equations \ref{equ2} \& \ref{equ3}, (\(\ast\)) represents the convolution operator, \(\sigma\) is the \textit{Leaky}ReLu activation function. The encoder network \(f_{\phi}\), consist of two convolutional layers on top of the input spectrogram. These convolutional layers are used for extracting hierarchical location invariant spatial features. The output of the last convolutional layer in \(f_{\phi}\) is then flattened and passed to a \(d\)-dimensional dense layer (\(h\)). This dense layer (\(h\)) is the embedding layer which learns the contextual representation of the spoken word corresponding to input \(W_t\) (contextualized on the context spoken word spectrograms). The decoder network takes the embedding layer (\(h\)) as input and generates a reconstruction \(W_t^r\) by passing (\(h\)) through a dense layer and two transpose convolutional layers. The \(d\)-dimensional embedding layer (\(h\)) learns an efficient contextualized representation of the word corresponding to \(W_t\) by minimizing the loss function \(L\) (shown in equation \ref{equ5}). In the equation below, \(N\) represents the batch size and \(m\) represents the size of the context window.
\begin{equation} \label{equ5}
\small{L_{\phi,\theta} = \frac{1}{N}\sum\limits_{t=1}^{N} (\frac{1}{2m}\sum\limits_{j=-m;j\neq0}^{m}(\Delta_{W_t,W_{t+j}}))}
\end{equation}
where,
\begin{equation} \label{equ5_1}
\small{\Delta_{W_t,W_{t+j}} = ||g_{\theta}(f_{\phi}(W_t))-W_{t+j}||_2^2}
\end{equation}

The lost function defined above helps the latent embedding to learn the contextual relationship between the target spoken word spectrograms and it's the corresponding context by calculating a reconstruction loss between the reconstruction \(W_t^r\) and the corresponding contextual spectrograms \(W_{t+j}\). Since a word spoken in different tones has different spectrograms, the model also captures the tone in which the words are uttered in its contextual embedding. So in summary the proposed model can not only incorporate context in the spoken word representations but can also incorporate its tone.

\section{Evaluation Setup}

\subsection{Dataset}
The proposed model uses Trump's speeches (Audio and word transcription)\footnote{https://www.kaggle.com/etaifour/trump-speeches-audio-and-word-transcription} dataset for training and testing. This dataset was chosen because it comprises of audio files and their corresponding word split JSON files. Another reason for choosing this dataset was that it contains speeches of a single person (which will eliminate the problem of having different pronunciations of the same word). These JSON files contain a direct mapping between each word spoken and the duration in which it was spoken. These mappings were used to split the full audio file into multiple audio files for each word spoken. This context mapping was used to create input-output pairs for the proposed model. The statistics about the dataset is shown in Table \ref{tab:my-data}.
\begin{table}[ht]
\centering
\caption{\small{Dataset Statistics}}
\label{tab:my-data}
\resizebox{0.9\columnwidth}{!}{%
\begin{tabular}{c|c|c|c}
\hlineB{3}
\# Words & \# Sentences & \# Context Mappings & \# Seconds \\ \hline \hline
 18.1k& 1K & 72.6K & 12.9K \\ \hlineB{3}
\end{tabular}%
}
\end{table}

\subsection{Training Details}
The proposed model was evaluated on 10\% of the data, and the rest was used for training. From the training set, 10\% data was used as the validation set. It was trained for 50 epochs having a mini-batch size of 5. For optimization, Adam optimizer was used to have an initial learning rate of \(0.01\). A context window size of 2 was used for all the experiments (due to computational resource limitations). Early stopping with the patience of 5 epochs and dropout with a dropout rate of \(0.7\) was used to avoid over-fitting. The size of the latent representation \(d\) was set to 16. The size of the filters in the convolutional and de-convolutional layers was set to (4$\times$4).

% Please add the following required packages to your document preamble:
% \usepackage{graphicx}

\subsection{Results}
The performance of the proposed model was validated by (1) inspecting the vector space and (2) evaluating its performance on three benchmark datasets for measuring word similarities, and comparing the proposed model with text-based language models (trained on the textual transcripts).

\begin{figure}[ht]
\centering
\includegraphics[width=0.8\linewidth]{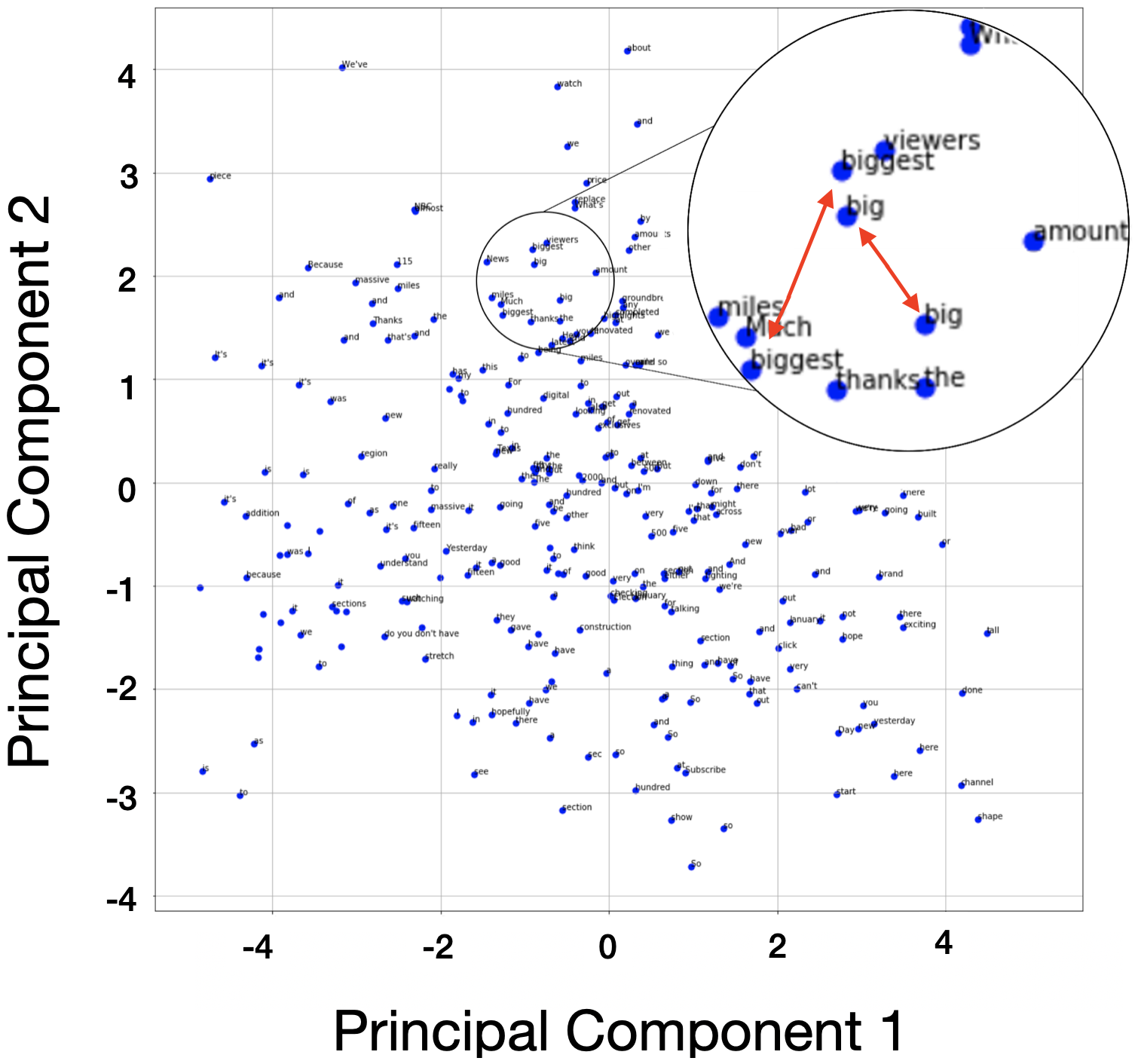}
\caption{\small{Vector space generated by the proposed model}}
\label{fig:model_pca}
\end{figure}

To visualize the performance of the proposed model, the dimensionality of the audio vectors (16-dimensional) was reduced using principal component analysis (PCA) \cite{hotelling1933analysis}, to plot the spoken word representations in a (2-dimensional) vectors space. Figure \ref{fig:model_pca} illustrates the vector space generated by the proposed model. On closeup, it can be easily seen that similar spoken words were grouped together in the vector space. For example, spoken words like \textit{big}, \textit{biggest}, and \textit{much} fall in vicinity to each other. It can also be seen in figure \ref{fig:model_pca} that the same spoken words (\textit{big} \& \textit{biggest}), uttered in different tones were also in close proximity but were slightly distant from each other. This demonstrates the capability of the model to capturing semantical and syntactical similarities between different spoken words (or the same spoken word in different tones).

\begin{table}[ht]
\centering
\caption{\small{Results Table}}
\label{tab:vector}
\resizebox{0.8\columnwidth}{!}{%
\begin{tabular}{c|c|ccc}
\hlineB{3}
\multirow{2}{*}{Dataset} & \# Word & \multicolumn{3}{c}{\small{Spearman’s rank correlation coefficient $\rho$}} \\ \cline{3-5} 
 & Pairs & \multicolumn{1}{c|}{Our Model} & \multicolumn{1}{c|}{Word2Vec} & GloVe \\ \hline \hline
SimVerb & 275 & 0.31 & \textbf{0.32} & 0.28 \\ \cline{1-1}
WS-SIM & 33 & \textbf{0.51} & 0.49 & 0.47 \\ \cline{1-1}
WS-REL & 53 & 0.23 & \textbf{0.25} & \textbf{0.25} \\ \hlineB{3}
\end{tabular}%
}
\end{table}

The spoken word representations generated by the proposed model were evaluated on three different benchmark datasets (\textbf{SimVerb} \cite{gerz-etal-2016-simverb}, \textbf{WS-SIM} and \textbf{WS-REL} \cite{agirre-etal-2009-study}) that are widely used for computing word similarities/relatedness between words. The comparison of the proposed model is done with the text-based language models Word2Vec \cite{NIPS2013_5021} and GloVe \cite{pennington-etal-2014-glove}. In the case of the proposed model, word similarities were obtained by measuring the cosine similarities between the spoken vector representations of the corresponding words, and in the case of Word2Vec \& GloVe,  similarities were computed between the corresponding word (textual) vector representations.  Table \ref{tab:vector} reports Spearman’s rank correlation coefficient $\rho$ between the human ranking \cite{myers2010research} and the ones generated by each model. The proposed model was trained on a small dataset (small vocabulary). So the proposed model was not able to generate representations for some of the word pairs present in the above mentioned three benchmark datasets (Number of word pairs used is also shown in the table above). Despite spoken words having different tones/expressing/pause for the same words depending on the context (in contrast to text), the proposed model was able perform comparably to the existing text-based language models.

\section{Conclusion}
This paper introduces an unsupervised model that not only was able to successfully generate semantically and syntactically accurate contextualized representations of varying length spoken words but was also able to perform adequately on three benchmark datasets for measuring word similarities. The proposed model also demonstrated its capabilities to capture tones and expressions of the spoken words. To the best of our knowledge, this is the first work that tries to model variable-length spoken words using convolutional autoencoders. In the future, we plan to extend the capabilities of the model to handle different pronunciations/accent by different speakers.

\bibliography{nlp4MusA}
\bibliographystyle{nlp4MusA_natbib}

\end{document}